\documentclass{article}

\usepackage[T1]{fontenc}
\usepackage{microtype}
\usepackage{graphicx}
\usepackage{subfig}
\usepackage{framed}
\usepackage{booktabs}
\usepackage{natbib}
\usepackage{amsmath}
\usepackage{amssymb}
\usepackage{IEEEtrantools}
\usepackage{adjustbox}
\usepackage{tikz}
\usetikzlibrary{positioning}
\usetikzlibrary[positioning, shapes, quotes, arrows, fit]
\tikzstyle{arrow} = [thick,->]
\tikzstyle{squarednode} = [rectangle, draw=black, thick]
\usetikzlibrary{shapes.multipart}
\usepackage{hyperref}

\usepackage[accepted]{icml2021}

\begin{document}

\twocolumn[
\icmltitle{Implicit Neural Representation
           For Accurate CFD Flow Field Prediction}

\begin{icmlauthorlist}
\icmlauthor{Laurent de Vito}{mtu}
\icmlauthor{Nils Pinnau}{microsoft,internship}
\icmlauthor{Simone Dey}{itestra,internship}
\end{icmlauthorlist}

\icmlaffiliation{mtu}{MTU Aero Engines AG, Munich, Germany}
\icmlaffiliation{microsoft}{Microsoft, Munich, Germany, nils.pinnau@live.de}
\icmlaffiliation{itestra}{itestra, Munich, Germany, simi.dey@gmx.de}
\icmlaffiliation{internship}{Work done during an internship at MTU Aero Engines AG, Munich, Germany}

\icmlcorrespondingauthor{Laurent de Vito}{laurent.vito@mtu.de}
\icmlkeywords{Machine Learning, ICML}

\vskip 0.3in
]

\printAffiliationsAndNotice{}

\begin{abstract}
Despite the plethora of deep learning frameworks for flow field prediction,
most of them deal with flow fields on regular domains, and
although the best ones can cope with irregular domains, they mostly rely on graph networks,
so that real industrial applications remain currently elusive.
We present a deep learning framework for 3D flow field prediction
applied to blades of aircraft engine turbines and compressors.  
Crucially, we view any 3D field as a function from coordinates that is modeled by a neural network we call the backbone-net.
It inherits the property of coordinate-based MLPs, namely the discretization-agnostic representation of flow fields in domains of arbitrary topology
at infinite resolution.
First, we demonstrate the performance of the backbone-net solo in regressing 3D steady simulations of single blade rows in various flow regimes:
it can accurately render important flow characteristics
such as  boundary layers,  wakes and shock waves.
Second, we introduce a hyper-net that maps the surface mesh of a blade
to the parameters of the backbone-net.
By doing so, the flow solution can be directly predicted from the blade geometry,
irrespective of its parameterization.
Together, backbone-net and hyper-net form a highly-accurate memory-efficient data-driven proxy to CFD solvers
with good generalization on unseen geometries.
\end{abstract}

\section{Introduction}
\label{Introduction}

There is a recent surge to devise fast deep learning models
as a substitute for Computational Fluid Dynamics (CFD) solvers
for the prediction of fluid flows~\cite{vinuesa2021potential}.
Indeed, CFD solvers are, though accurate, particularly slow, and this problem
gets compounded in design optimization \cite{popov2020optimization},
possibly under manufacturing uncertainties \cite{kamenik2018robust,meyer2019aerodynamic},
through numerous calls to the flow solver.

Over few years, we have witnessed CNN-based models, followed by graph-based models
and point cloud models for fluid flow prediction.
It is worth noticing that those models are not intrusive.
This is a key property for
their wide-spread adoption in industry. They are data-driven:
They consume data produced by expensive CFD solvers for training. 

It is natural to re-interpret nodes of a computational mesh as pixels,
and make use of Convolutional Neural Networks (CNNs)
to solve expensive equations~\cite{xiao2018novel,gao2021super}
but this approach is clearly limited to equally-spaced Cartesian meshes (a regular grid in academic use only).
Unstructured meshes\footnote{We regard body-fitted multi-block meshes as unstructured meshes
in opposition to equally-spaced Cartesian meshes.}
are widely used in the industry because they can adequately delineate complex geometries
and easily deal with localized regions that require different resolution, e.g.
cells are small near walls to capture the boundary layer where flow transitions are sharp
whereas they are large in the freestream where gradients are smooth.
To make unstructured meshes amenable to a treatment with CNNs, they are rasterized
into voxel grids and processed
using 3D volumetric convolutions~\cite{thuerey2020deep,aulich2019surrogate}.
With low resolutions, some information is inevitably lost during voxelization.
Therefore, a high voxel resolution is required to preserve flow details.
However, scalability is poor because the computational cost and memory
requirement both increase cubically with voxel resolution.
Thus, it is infeasible to train a voxel-based model with high-resolution grids.

A mesh can equivalently be viewed as a graph.
So Graph Neural Networks (GNNs) are legitimate candidates
for predicting flow fields~\cite{pfaff2020learning,meyer2021deep,baque2018geodesic}.
Since they rely on message-passing, propagating the information from one node
to a distant one requires stacking many graph convolutional layers.
Pooling is an effective technique to increase the receptive field of nodes
but what is a cheap operation in CNNs turns out to be challenging in GNNs~\cite{grattarola2021understanding}.

If the adjacency matrix of the graph is discarded, we end up with a point cloud.
Many point cloud techniques are in fact graph techniques in disguise: A
graph is built by linking a node to its nearest neighbors.
Point cloud models for flow field prediction are uncommon~\cite{kashefi2021point}.

Our work departs radically from those approaches that only yield 
a discrete representation of the flow field.
Instead, we rely on an \emph{implicit neural representation} of the flow field:
A flow field  is viewed as a function from coordinates
that is approximated by a neural network. Given an input coordinate,
the network is trained to output the value of the flow field at that coordinate.
Such neural networks are commonly referred to as
coordinate-based networks or simply coordinate networks in the literature.
They have become popular since~\cite{tancik2020fourier}
showed how to enable a neural network to learn high frequency functions
in low-dimensional problems, typically in 2D and 3D. 

Once trained on a simulation,
our coordinate MLP,  called here the \emph{backbone-net}, offers
a compressed mesh-agnostic representation of the flow field it was trained on.
Indeed, our backbone-net is small, so its few weights
can be stored instead of the full CFD solution.
But apart for data reduction \cite{zhang2021implicit,huang2022compressing},
this is not particularly useful in itself.

In this work, we are concerned with steady CFD flow fields
of single 3D blade rows of
aircraft engine turbines and compressors.
We would like to capture the change in flow solution when the
blade geometry is varied. Since the flow solution
is equivalently viewed as the weights of the backbone-net,
we would like to adapt those weights with the geometry of the blade.
This task is per definition devoted to a hyper-network \cite{ha2016hypernetworks}.
Our hypernetwork, here in short \emph{hyper-net}, takes in the blade
geometry in the form of a triangulated surface mesh, and yields the weights of the backbone-net.
Our solution is thus independent of the blade parameterization.
This makes our model even more appealing as a proxy for a CFD solver in an optimization:
we can use the same model for different optimizations
even if the degrees of freedom of the parameterization change.
But there is another advantage in taking the surface mesh as input to our model:
some geometrical features, e.g. fillets, small gaps or steps, 
are hard to describe in a compact form that can be passed to the model.
Because our model ingests the surface mesh of the configuration,
all those geometrical features are no longer obstacles to
a faithful representation of the configurations.
All in all, we substitute an optimized CFD flow solver by a single forward pass of a hyper-net.
Whereas the backbone-net is trained on a single simulation,
the hyper-net is trained on a set of simulations.
The hyper-net is end-to-end differentiable and so it is trained
using stochastic gradient descent.

The contributions of this paper are as follows:
\begin{itemize}
\itemsep0em
\item We introduce an implicit neural representation for CFD flow fields in turbomachines.
As a result, our representation of flow fields
on 3D unstructured meshes is compact and continuous.
The advantage is threefold: (1) This representation
does not change if the mesh is modified 
as long as the flow field is the same\footnote{
For this property to hold, the flow is in the asymptotic regime
and the mesh is modified 
keeping the mesh resolution unchanged.};
(2) The interpolation to unseen coordinates is smooth;
(3) Because we can query the flow field at any coordinates,
the flow field has virtually an infinite resolution,
a property that was leveraged in image super-resolution~\cite{klocek2019hypernetwork}.

\item Our method does not require extra preprocessing like the time-consuming and approximate interpolation of 
CFD flow fields onto a Cartesian mesh.

\item We demonstrate that our backbone-net, a small coordinate MLP,
with discrete Fourier features,
can render the full 3D flow field of turbomachine configurations in various flow regimes,
from subsonic to supersonic, faithfully.
In \cite{white2020fast}, a mixture of experts was advocated where each mixture is 
a simple MLP. The rational behind this choice was that the flow solution is too complex
to be accurately predicted by a single MLP. We challenge this claim and
show that a coordinate MLP with a carefully tailored architecture
can adequately predict CFD flow fields. By doing so,
we circumvent the difficult problem of initializing mixture of experts~\cite{makkuva2020learning}.

\item We establish the direct mapping of the blade geometry to the aerodynamic flow fields
using a hypernetwork. The idea of using a hypernetwork is attractive
because it can more generally modulate the (weights of) backbone-net based
on any relevant information.
In this work, we condition the flow solution on the blade geometry,
but our framework can incorporate any
other type of side information, e.g. boundary conditions.

\item We empirically show that our model generalizes well from few samples.
\end{itemize}

We focus here exclusively on flow solutions of compressor and turbine blades,
but this concept is not limited to internal aerodynamics:
it can be applied more broadly to any solution of
systems of partial differential equations~\cite{pan2022neural}.

\section{Related Work}

Our work is closely related to \cite{pan2022neural}:
In their work, a hyper-net they call ParameterNet produces the weights and biases of 
a backbone-net they call ShapeNet based on external factors.
ShapeNet is also a coordinate-based MLP.
They showed the benefits of this new paradigm
in a broad range of applications notably in CFD.
Furthermore, they compared this approach against other recent frameworks like DeepONet.

In this work, we leverage a similar framework 
and apply it in a real-world large-scale industrial setting.
Whereas in \cite{pan2022neural} the input to their ParameterNet
is a time stamp because their focus is on unsteady simulations,
the input to our hyper-net is a blade geometry in the form of a surface mesh.
Furthermore, their ShapeNet has sine activation functions \cite{sitzmann2020implicit}, 
but our backbone-net is equipped with deterministic Fourier features because we found sine activation functions
to underperform.

\textbf{Convolutional Neural Networks (CNNs).}
Traditional CNNs can be applied
to predict the flow solution on regular grids, either in 2D or 3D 
\cite{guo2016convolutional,aulich2019surrogate,thuerey2020deep,obiols2020cfdnet,chen2021numerical},
as well as on irregular structured grids by mapping them to canonical regular grids \cite{chen2021towards}.
However, CNNs do not scale well because computational and memory requirements
grow cubicly with the 3D grid resolution. As a result, CNNs cannot
exploit the detailed geometry of irregular bodies in fluids
and cannot render important features such as boundary layers, wakes and shock waves.

Compared to CNNs, our model has a very low memory footprint,
so the infrastructure cost for training and deploying our model is reduced.
Furthermore, it faithfully predicts important flow features.

\textbf{Graph Neural Networks (GNN).}
Opposite to CNNS, GNNs have the potential to work natively on arbitrary meshes.
Impressive results haven been obtained recently
with GNN-based methods that directly mimic the CFD solver iterative process in computing
the solution on a mesh from one time-step to the next \cite{pfaff2020learning,meyer2021deep}.
However, GNNs scale badly with the mesh size.
The problem is that neighboring nodes affect each other in the learning process.
For the computation of the embedding of a single node, a GNN needs the
embeddings of the neighbors of this node. Consequently, GNNs must learn on the full graph,
which is infeasible if the graph is too large.
Mini-batching is a solution,
but despite advances \cite{ding2021vq,klicpera2021locality},
training graph neural networks on large graphs is still challenging.
By avoiding graph neural networks, our code is considerably simpler (no need to introduce edges)
and more efficient (no information gathering at each node from neighbors)
than \cite{pfaff2020learning}.
\cite{meyer2021deep} proposed a data-driven proxy that necessitates a hierarchy of meshes,
which makes it cumbersome to code and certainly precludes its extension to 3D.

The output of GNNs is furthermore inherently discrete: the solution is
available only at the nodes of the mesh.
With our method, the solution can be queried anywhere in space (and time if
time is included as input feature).

Furthermore, most graph-based methods are confined to 2D configurations
in subsonic flow regime \cite{bonnet2022airfrans}.
We consider 3D configurations in
subsonic up to supersonic flow regimes.

\textbf{Hypernetworks.} Hypernetworks are models
that generate parameters for other models \cite{ha2016hypernetworks}.
Many recent works rely on hypernetworks
\cite{kang2017incorporating,sitzmann2019scene,littwin2019deep,spurek2020hypernetwork,knyazev2021parameter,skorokhodov2021adversarial}.
This success is rooted in the modularity property of hypernetworks \cite{galanti2020modularity}.
That hypernetworks outperform embedding-based methods was experimentally illustrated
in \cite{skorokhodov2021adversarial}.

\textbf{Physic-Informed Neural Networks (PINNs).} 
The models mentioned above are
proxies (also called deep surrogates)
to CFD solvers that need training data.
On the opposite, physic-informed neural networks~\cite{raissi2019physics}
solve for the motion of fluid flows just as CFD solvers do.
Hence, their generalization power is on par with that of CFD solvers.
This property makes them highly attractive, but despite all the hype, they suffer from severe deficiencies:
1/ They have not matured yet
as a drop-in replacement for CFD solvers
as they currently cannot compete with advanced PDE solvers in terms of accuracy
\cite{wang2021understanding,chuang2022experience}
2/ The laborious effort to encode all the equations to solve, including turbulent and transition models,
and the various boundary conditions, is overlooked, though it is a huge undertaking \cite{du2022autoke};
3/ Mass, momentum and energy are not conserved, because the solution is solved point-wise,
whereas  finite volume flow solvers have the mass, momentum and energy conservation property built-in. 
Those drawbacks hamper the wide-spread use of PINNs.

\textbf{Implicit Neural Representation (INR).}
In implicit neural representation, a discrete signal,
e.g. the color information on a lattice (an image) or on a low-dimensional manifold (a meshed shape),
is represented as a continuous function by a neural network,
usually a MLP with the ReLU activation function.
Since the input to those networks are the low-dimensional coordinates,
those networks are referred to as coordinate-based MLPs.
Not only is the representation continuous but also compact.
Coordinate-based MLPs have been used to represent images \cite{stanley2007compositional},
volume density \cite{mildenhall2020nerf},
occupancy \cite{mescheder2019occupancy},
signed distance \cite{park2019deepsdf}
and have been employed in a variety of other tasks.
Coordinate MLPs have difficulty learning high frequency functions \cite{rahaman2019spectral}.
To overcome this limitation, \cite{mildenhall2020nerf} 
propose to map the raw coordinates to deterministic Fourier features
whereas in \cite{tancik2020fourier} 
they are mapped to random Fourier features.
\cite{sitzmann2019scene} take another approach and use sinusoidal activation functions.

\section{Models}

\begin{figure*}[ht!]
\centering
\begin{tikzpicture}[invisiblenode/.style={rectangle, draw=white!60, fill=white!5, ultra thin, minimum size=0mm}]
\node (theta) at (3,0) {$\theta$};
\node[inner sep=0pt] (blade_mesh) at (8,0)
    {\scalebox{0.9}[0.9]{\includegraphics[width=.1\textwidth,trim=.1cm .1cm .1cm .1cm, clip]{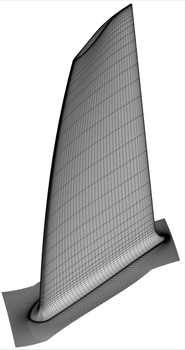}}};
\node[inner sep=0pt] (blade_value) at (12,0)
    {\scalebox{0.9}[0.9]{\includegraphics[width=.1\textwidth,trim=.1cm .1cm .1cm .1cm, clip]{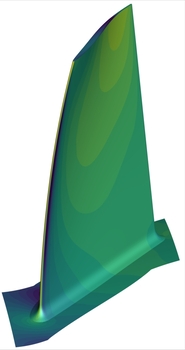}}};
\node (q) at (16,0) {$q$};
\draw[->,thick] (theta) -- node [midway,above=0em ] {Mesh generator} (blade_mesh);
\draw[->,thick] ([yshift=+0ex]blade_mesh.east) -- node [midway,above=0em ] {Flow solver} ([yshift=+0ex]blade_value.west);
\draw[->,thick,red] ([yshift=-9ex]blade_mesh.east) -- node [midway,above=0em ] {Backbone-Net} ([yshift=-9ex]blade_value.west);
\draw[->,thick] (blade_value) -- node [midway,above=0em ] {Postprocessor} (q);

\draw[->,thick,white]
   ([yshift=-21ex]blade_mesh.east)
   -- node [text=red,midway,above=-1.5em ] {
        $\text{NN}_\phi: \boldsymbol{x} = \begin{bmatrix}x\\y\\z\end{bmatrix} \rightarrow \hat{\boldsymbol{y}} = \begin{bmatrix}\rho\\p\\V_x\\V_y\\V_z\end{bmatrix}$}
    ([yshift=-21ex]blade_value.west);

\node[inner sep=0pt] (blade_mesh2) at (0,-4)
    {\scalebox{0.9}[0.9]{\includegraphics[width=.1\textwidth,trim=.1cm .1cm .1cm .1cm, clip]{./images/img_0.jpg}}};
\node (theta_prime) at (3,-4.) [text=red] {$\theta'$};
\node (phi) at ( theta_prime -| blade_mesh) [text=red] {$\phi$};
\node[below = 1.7cm of blade_mesh] (below_NN_phi) {};

\draw[->,thick, red] (blade_mesh2) -- node [midway,above=0em ] {GNN} (theta_prime);
\draw[->,thick, red] ([yshift=+0ex]theta_prime.east) -- node [midway,above=0em ] {Hyper-Net} ([yshift=+0ex]phi.west);
\draw[->,thick,white]
   ([yshift=-1ex]theta_prime.east)
   -- node [text=red,midway,above=-1.5em ] {
        $\text{NN}_\psi: \theta' \rightarrow \phi$}
    ([yshift=-1ex]phi.west);
\draw[->,thick, red] (phi.north) -- (below_NN_phi.center);

\end{tikzpicture}
\vskip -0.1in
\caption{Overview of the backbone-net and hyper-net. At the top in black, we have the usual tool chain: From the blade design vector $\theta$,
a 3D mesh is generated (only the surface mesh is depicted), then the CFD solver computes the full 3D flow solution and eventually the postprocessor
aggregates the local information at the cell-centers and outputs quantities of interest denoted by $q$ like massflow and efficiency. At the
bottom in red, we have our model that comprises the backbone-net and the hyper-net.
The backbone-net maps the coordinates of a point of the computational domain to its flow features.
It is specialized to a configuration.
We make it capable of dealing with any blade by predicting its weights and biases
from the surface mesh of a blade given as input. First, a graph neural network (GNN) extracts
a fixed-sized vector representation of the blade, $\theta'$. This representation can be thought of as 
a pseudo design vector.
Afterwards, the hyper-net generates the weights and biases, $\phi$, of the backbone-net.
Finally, the backbone-net yields the 3D flow solution that is compared to the ground truth.}
\label{fig:overview}
\vskip -0.15in
\end{figure*}

A overview of the full model is given in Figure~\ref{fig:overview}.
All models are implemented in Python using PyTorch \cite{paszke2017automatic}.
We emphasize that our implementation is simpler compared to other approaches:
The hyper-net is a small  residual networks \cite{he2016deep}
and the backbone-net is a fully-connected network,
both do not require any specialized modules and so
are easily coded using any deep learning framework.

\tikzstyle{line} = [draw]

\subsection{The Backbone-Net}

The backbone-net is a small coordinate-based MLP that takes in a coordinate vector $\boldsymbol{x}$ and
outputs an approximation $\hat{\boldsymbol{y}}$ to the flow features $\boldsymbol{y}$ at that position.
Typically, the geometry of a blade is specified by the design vector $\boldsymbol{\theta}$.
From the design vector, the 3D computational domain around the blade is defined and  meshed, and
the flow solution for prescribed boundary conditions
is computed at cell-centers by the flow solver. So the dataset for training the backbone-net is 
$\mathcal{D}^{b-net} = \{ (\boldsymbol{x}_n,\boldsymbol{y}_n) \}_{n=1}^N$
with cell-center coordinates $\boldsymbol{x}_n \in \mathbb{R}^{D_x}$
and flow features $\boldsymbol{y}_n \in \mathbb{R}^{D_y}$
for $n=1,\cdots,N$ given by the flow solver.
It is clear that the nature of the mesh --- block-structured, unstructured or hybrid --- is irrelevant.

The input $\boldsymbol{x}$ is optionally fed into a positional encoder $\text{PE}$
as introduced in \cite{tancik2020fourier} and the result is linearly transformed after concatenation with the input:
$\boldsymbol{z}_0^{(b)} = \boldsymbol{W}_{in}^{(b)} [ \text{PE}(\boldsymbol{x}) \Vert \boldsymbol{x}]$.
We omit the bias term of the linear transformation for sake of simplicity.
$\Vert$ denotes concatenation.
Concatenation of raw data $\boldsymbol{x}$ with Fourier features $\text{PE}(\boldsymbol{x})$
was found beneficial as in \cite{chen2019learning,jaegle2021perceiver}.
Opposite to \cite{tancik2020fourier},
the encoder does not construct random Fourier features.
Following \cite{mildenhall2020nerf},
the positional encoding scheme is deterministic: The mapping contains
only on-axis frequencies.
The positional encoder has base frequency $f_b$
and the $L$ on-axis frequencies are defined as $f_l = 2 \pi (2^l f_b)$ for $l=0,\cdots,L-1$ \cite{mildenhall2020nerf}.
We experimentally observed that the benefit of a positional encoding with $L > 4$ is negligible though it does not hurt performance.

Following the positional encoder is a stack of fully connected layers with identical layout:
\begin{equation}
\boldsymbol{z}_{k}^{(b)} = \sigma( \boldsymbol{W}_{k}^{(b)} \boldsymbol{z}_{k-1}^{(b)} )
, \ \ \
k=1,\cdots,K
\label{eq:fcblock}
\end{equation}
$\sigma$ designates the GELU activation function \cite{hendrycks2016gaussian}.
We finally insert a linear layer to convert to the expected output:
$\hat{\boldsymbol{y}} = \boldsymbol{W}_{out}^{(b)} \boldsymbol{z}_K^{(b)}$.
In all experiments, the positional encoder has base frequency $f_b=0.5$ 
and the number of Fourier features is $L=4$.
The backbone-net has $K=6$ layers and the hidden dimension is 112.

After the backbone-net is trained on the dataset $\mathcal{D}^{b-net}$ in a supervised manner,
its weights are implicitly dependent on the design vector $\boldsymbol{\theta}$ that
specifies the geometry of the blade for which we have the simulation.
We would like to make this dependency explicit.
Stated otherwise, we would like to directly get the flow solution
(or equivalently the weights of the backbone-net denoted collectively by $\phi$)
as a function of the design vector, or, more generally,
as a function of the blade geometry.
For that purpose, we introduce a hyper-net that
predicts the weights of the backbone-net from the blade geometry.
By doing so, the expensive flow solver is replaced with a single forward pass of the hyper-net.

\subsection{The Hyper-Net}
\label{The Hyper-Net}

Since the hyper-net yields the weights of another neural network, namely the backbone-net,
the dataset for training the hyper-net is a dataset of simulations (configurations):
$\mathcal{D}^{h-net} = \{ (\mathcal{G}_m, \mathcal{D}^{b-net}_m) \}_{m=1}^M$.
$\mathcal{G}_m$ designates the geometry of the $m$-th blade in the form of a triangulated surface.
It is equivalent to a graph.
Each $\mathcal{D}^{b-net}_m = \{ ( \boldsymbol{x}_{m,n}, \boldsymbol{y}_{m,n}) \}_{n=1}^{N_m}$ is the 3D flow solution
for the blade geometry $\mathcal{G}_m$.
The number of points $N_m$ in each of those simulations can vary
but they are the same in our experiments.

The input to the hyper-net is a fixed-size pseudo-design vector (or blade embedding), $\boldsymbol{\theta}'$, representing the geometry.
It is obtained by passing the geometry through a graph neural network (GNN).
Typically, a GNN uses a stack of message passing layers~\cite{gilmer2017neural}
to learn powerful embeddings compared to a plain MLP. However, 
they are computationally expensive to train. We experimentally 
found out that we can achieve comparable performance using the simple and much faster
PointNet architecture~\cite{qi2017pointnet}. It is a MLP followed by global max pooling.
This suggests that, at least for the task of predicting the CFD flow fields from the blade geometry
(or its embedding), a complex architecture is not crucial.
The input, a triangulated surface mesh, is thus treated as a point cloud.
This is fortunate as the triangulation of the original surface mesh
is not unique.
In our experiments, the MLP is a residual network with 4 layers,
a hidden dimension of 128 in the main branch and 192 in the residual branch
with GELU nonlinearities.
The pseudo design vector has dimension 16 which is surely greater than the
intrinsic dimension of the datasets we consider experimentally.
Indeed, the blade parameterization has
at most 17 degrees of freedom, see Tables \ref{table:hyper-summary-comp} and \ref{table:hyper-summary-turb},
and reconstruction of blades is 
as good as with pseudo design vectors of greater size, see Figure \ref{fig:turb_stator_geo}.

The hyper-net takes in the blade embedding $\boldsymbol{\theta}'$
and outputs the weights $\phi$ of the backbone-net.
The hyper-net is a small residual network:
\begin{IEEEeqnarray}{ccl}
\boldsymbol{z}_{0}^{(h)} &=& \boldsymbol{W}_{in}^{(h)} \boldsymbol{\theta} \\
\boldsymbol{z}_{k}^{(h)} &=& 
\boldsymbol{z}_{k-1}^{(h)}
+
\boldsymbol{W}_{2,k}^{(h)} \sigma( \boldsymbol{W}_{1,k}^{(h)} \sigma(\boldsymbol{z}_{k-1}^{(h)}) ),
1 \le k \le K'
\\
\phi &=& \boldsymbol{W}_{out}^{(h)} \boldsymbol{z}_{K'}^{(h)}
\label{eq:hypernet_arch}
\end{IEEEeqnarray}
Notice that we insert a GELU non-linearity right at the start of the residual branch.
Experimentally the hyper-net is shallow  with $K' = 1$ residual blocks;
in each block, the hidden dimension of the main branch is 48
while it is 96 in the residual branch.
In the residual branch dropout is enabled to provide additional regularization in the small data regime.

Training a hypernetwork is notoriously difficult \cite{lorraine2018stochastic,ukai2018hypernetwork}.
We follow \cite{ortiz2023magnitude} and treat the hyper-net predictions
as additive changes to the backbone-net. This makes training stable.

Hypernetworks are also prohibitively expensive, in both compute and memory.
To avoid an explosion of the number of parameters,
only the biases of the backbone-net are predicted by the hyper-net.
This is similar to applying shift modulations~\cite{naour2023time}.
However, by doing so, the performance of our model  degrades markedly.
To counter-act this loss of performance, the hyper-net also predicts the first and the last weight
matrices of the backbone-net, $\boldsymbol{W}_{in}^{(b)}$ and $\boldsymbol{W}_{out}^{(b)}$ respectively.
The hyper-net is thus small, with 260k parameters in our experiments.

\section{Training}

We train the backbone-net solo for 300 epochs and the hyper-net for 400 epochs
using NAdam \cite{dozat2016incorporating}
and a cosine learning rate scheduler.
The initial learning rate is 0.01 for the backbone-net solo and 0.001 for the hyper-net.

The backbone-net is small and data is plenty, so it is not regularized.
But the hyper-net is trained with limited data, so 
it is regularized with dropout.
The dropout rate is a hyperparameter that is found by grid search using Optuna \cite{optuna_2019}.

We train the backbone-net solo with mini-batches.
With large batch sizes generalization is poor, in accordance
with \cite{keskar2016large} that showed that large batch sizes 
are associated to a degradation in model quality,
whereas with small batch sizes
the training run-time takes longer.
We set the trade-off by a batch size of 0.5k.
Training takes circa 20 minutes on a single core
of an Intel Xeon Platinum CPU
for a dataset with around 0.4Mio points.
We also train the hyper-net with mini-batches:
each batch has 20k points of a single configuration
associated to a given blade geometry.
Notice that we consider all points of a configuration at each epoch.
The training takes less than one day on 32 cores of a CPU
for a dataset with circa 130 configurations.

\section{Experiments}

In our experiments, both qualitative  and quantitative evaluations are provided.
Visual examination of 3D flow fields is hard,
so to avoid cluttered plots we only show slices at constant relative radius $r_{S_1}$,
ranging from 0 (hub) to 1 (tip). $r_{S_1} = 0.5$ is referred to as midspan.
See Figure \ref{fig:hypernet_comp_stator_full} for an illustration.

\subsection{Datasets}

We consider four different blades, a stator and a rotor from a low-pressure subsonic turbine,
\texttt{turb-stator} and \texttt{turb-rotor},
and a stator and a rotor from a low-pressure transonic compressor,
\texttt{comp-stator} and \texttt{comp-rotor}.
Those blades feature complex geometries:
Both turbine blades have large fillets,
the compressor stator has half-gaps
(small gaps in the rear part at the hub and tip, see Figure \ref{fig:hypernet_comp_stator_full}) and
the compressor rotor has a tip clearance (small gap between blade and casing).
For each of these four original blades indexed by $b$,
we sample its geometry vector $\boldsymbol{\theta}_b$ randomly $M_b$ times
by varying radially among other things the leading edge blade angle, trailing edge blade angle and stagger angle,
so as to construct a dataset of configurations $\mathcal{D}^{h-net}_b = \{ (\mathcal{G}_{b,m}, \mathcal{D}^{b-net}_{b,m}) \}_{m=1}^{M_b}$
for training the hyper-net.
As a result, all datasets $\mathcal{D}^{b-net}_{b,m}$ of a given series $b$ for $m=1,\cdots,M_b$
have the same number of points, approximately 0.4Mio.
Recall that $\mathcal{D}^{b-net}_{b,m}$ is the dataset from a simulation. 
The boundary conditions at inlet (total pressure, total temperature and velocity directions)
and outlet (back pressure), along with the rotational speed, are hold constant
for each series of blades.

All 3D steady compressible flow solutions $\mathcal{D}^{b-net}_{b,m}$ were generated using
the Navier-Stokes solver TRACE\footnote{The non-commercial TRACE solver is developed jointly by DLR and MTU.}
\cite{becker2010recent}
with the Wilcox $k-\omega$ turbulence model \cite{wilcox1988reassessment}.
For turbine blades, the $\gamma-Re_\theta$ transition model \cite{langtry2009correlation} was enabled.

Our model outputs $\hat{y} = [\rho,p,V_x,V_y,V_z]$. To those five primitive variables,
it is possible to add quantities related to the turbulence model
like the (log of the) turbulent viscosity,
but ultimately we are interest in quantities like massflow, efficiency and flow turning,
whose derivation relies only on the five primitive variables our model predicts.

All input and output features are linearly scaled into $[-1,1]$ based on the 
extreme values computed from samples in the training set only.

\subsection{Data Augmentations}

The stators and rotors are extracted from axial flow turbomachines.
So the flow solution is equivariant under a translation along the $x$-axis.
Furthermore, the flow solution of a configuration is the same if the configuration is rotated about the $x$-axis,
under the condition that the velocity components $V_y$ and $V_z$ are rotated accordingly.
A blade has no canonical position in space we could set it in.
Hence our model would fail to generalize to configurations in unknown positions
and experience a performance drop at test time.
To overcome this difficulty, we let our model learn those properties using data augmentations
(as is usually done in image classification \cite{alex1imagenet}).
We re-center each configuration at $x=0$ (by translation) and $y=0$ (by rotation) --- thereby defining approximately a reference position --
and at training time we apply a small random rotation (between -5° and +5° in all our experiments) about the $x$-axis to the configurations.
Since the extension of each configuration (of a given series) in the $x$ direction is the same, it is not necessary
to apply a random translation along the $x$-axis.
At test time, we simply put the configurations in the reference position.
This data augmentation mitigates overfitting
and makes the hyper-net robust against a departure from the reference position at inference.

\subsection{Evaluation Metrics}

\begin{figure}[ht]
\begin{center}
\scalebox{0.9}[0.7]{
\centerline{\includegraphics[width=\columnwidth, trim=0.cm 0.cm 0.cm 0.cm,clip]{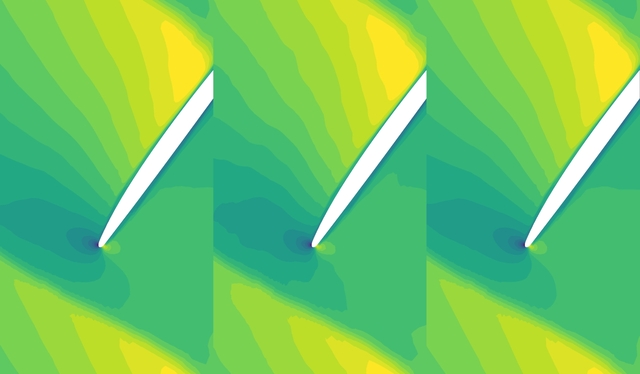}} 
}
\vskip 0.01in
\scalebox{0.9}[0.7]{
\centerline{\includegraphics[width=\columnwidth, trim=0.cm 0.cm 0.cm 0.cm,clip]{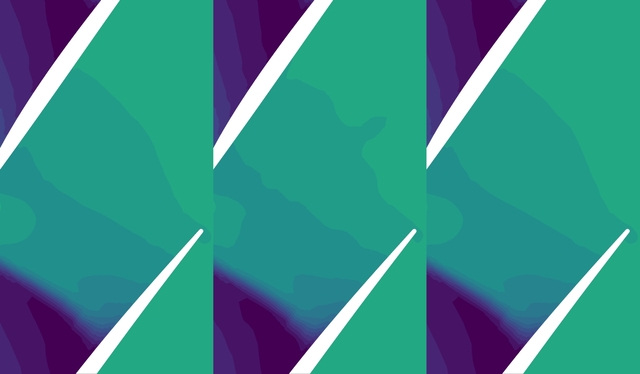}} 
}
\vskip -0.1in
\caption{Left: Predictions with MAE loss; middle: predictions with MSE loss; right: ground-truth, for the backbone-net trained solo on \texttt{comp-rotor}.
We show the axial velocity component $V_x$ at the leading edge (top row)
and the pressure $p$ at the trailing edge (bottom row), both at $r_{S_1} = 0.8$.}
\label{fig:backbone_solo_MAE_MSE_GT}
\end{center}
\end{figure}

In regression the choice of a differentiable cost function usually defaults to the mean squared error (MSE)
but we use the mean absolute error (MAE) as evaluation metric
since MAE produces results that are visually better,
as is the case in many image application tasks \cite{zhao2016loss,isola2017image}.
Figure~\ref{fig:backbone_solo_MAE_MSE_GT} illustrates the advantage of MAE over MSE in predicting flow solutions. 

\subsection{Backbone-Net Solo}

We check the capability of the backbone-net 
by training it solo
on various configurations.
For each configuration, we choose a 80/20 train/test split.
The training set is further divided into training and validation sets (90/10 split).
A summary of our results for the compressor and turbine configurations is presented
in Tables \ref{table:comp-backbone-summary} and \ref{table:turb-backbone-summary}, respectively.

\begin{figure}[!ht]
\begin{center}
\scalebox{0.9}[0.9]{
\centerline{\includegraphics[width=\columnwidth, trim=0.cm 0.5cm 0.cm 0.5cm,clip]{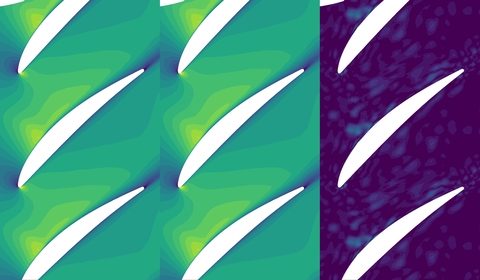}} 
}
\vskip 0.01in
\scalebox{0.9}[0.9]{
\centerline{\includegraphics[width=\columnwidth, trim=0.cm 0.5cm 0.cm 0.5cm,clip]{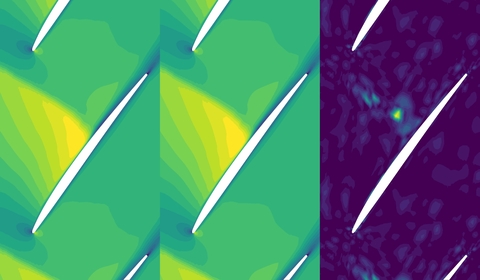}} 
}
\vskip -0.1in
\caption{Predictions of the axial velocity component $V_x$ for the configuration \texttt{comp-rotor} using the backbone trained solo (left) versus  ground-truth (middle);
absolute differences on the right.
Top row: Near hub ($r_{S_1}=0.1$).
Bottom row: Near tip ($r_{S_1}=0.8$).}
\label{fig:backbone_solo_comp_rotor}
\end{center}
\end{figure}

\begin{figure}[!ht]
\begin{center}
\scalebox{0.9}[0.6]{
\centerline{\includegraphics[width=\columnwidth, trim=0.cm 0.3cm 0.cm 0.4cm,clip]{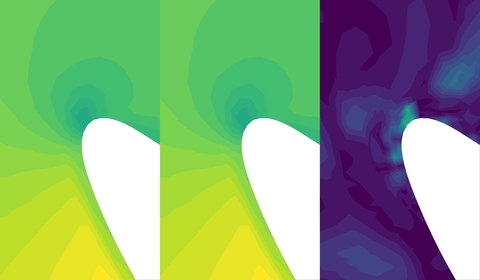}} 
}
\vskip 0.01in
\scalebox{0.9}[0.6]{
\centerline{\includegraphics[width=\columnwidth, trim=0.cm 0.3cm 0.cm 0.4cm,clip]{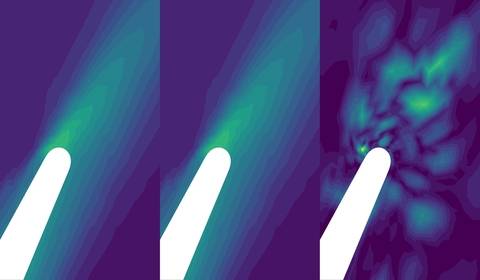}} 
}
\vskip -0.1in
\caption{Predictions of the circumferential velocity component $V_\theta$ for the configuration \texttt{turb-rotor}
using the backbone trained solo (left) versus  ground-truth (middle) at midspan;
absolute differences on the right.
Top row: at leading edge; bottom row: at trailing edge.}
\label{fig:backbone_solo_comp_rotor_LE_TE}
\end{center}
\end{figure}

Our backbone-net is rather small, with roughly 80k parameters.
It would be straightforward to scale up this model 
to get much better results
than the ones we will present in this section
without severely overfitting the training dataset
because a training dataset has circa 0.4Mio points.
However, we must keep in mind that
all the weights of the backbone-net will be later predicted by the hyper-net
and so we must refrain from building a too large backbone-net,
otherwise the hyper-net would require too many parameters.
Though small, the backbone-net is expressive enough.
Figure~\ref{fig:backbone_solo_comp_rotor} clearly illustrates its outstanding ability
in accurately predicting the flow fields of the primitive variables.
The flow is transonic close to the hub and supersonic close to the tip,
exhibiting a strong shock wave.
Differences against the ground-truth are barely noticeable visually.
Figure~\ref{fig:backbone_solo_comp_rotor_LE_TE} shows the ability of the backbone-net
in rendering the complex flow solution close to  leading and trailing edges.

\begin{table}[!t]
\caption{Summary of backbone-net solo training on the compressor datasets. A dataset is a simulation.}
\label{table:comp-backbone-summary}
\begin{center}
\vskip -0.2in
\begin{small}
\begin{sc}
\begin{tabular}{lrrr}
\toprule
Dataset & $\mathcal{D}^{b-net}_\text{comp-stator}$ & $\mathcal{D}^{b-net}_\text{comp-rotor}$ \\
\midrule
Training loss              & 4.7e-4 & 7.4e-4 & \\
Validation loss            & 5.3e-4 & 8.6e-4 & \\
Test loss                  & 5.2e-4 & 8.5e-4 & \\
\#Training samples (k)     & 416    & 307    & \\
\#Test samples (k)         & 104    & 77     & \\
\#Epochs                   & 300    & 300    & \\
Training time (minutes)    & 23     & 19     & \\
\bottomrule
\end{tabular}
\end{sc}
\end{small}
\end{center}
\vskip -0.15in
\end{table}

\begin{table}[!t]
\caption{Summary of backbone-net solo training on the turbine datasets. A dataset is a simulation.}
\label{table:turb-backbone-summary}
\begin{center}
\vskip -0.15in
\begin{small}
\begin{sc}
\begin{tabular}{lrrr}
\toprule
Dataset & $\mathcal{D}^{b-net}_\text{turb-stator}$ & $\mathcal{D}^{b-net}_\text{turb-rotor}$ \\
\midrule
Training loss              & 5.8e-4 & 6.7e-4 & \\
Validation loss            & 6.4e-4 & 7.4e-4 & \\
Test loss                  & 6.7e-4 & 7.4e-4 & \\
\#Training samples (k)     & 346    & 346    & \\
\#Test samples (k)         & 86     & 86     & \\
\#Epochs                   & 300    & 300    & \\
Training time (minutes)    & 19     & 19     & \\
\bottomrule
\end{tabular}
\end{sc}
\end{small}
\end{center}
\vskip -0.15in
\end{table}

\begin{figure}[!ht]
\begin{center}
\scalebox{1}[0.9]{
\centerline{\includegraphics[width=\columnwidth, trim=0.cm 0.3cm 0.cm 0.4cm,clip]{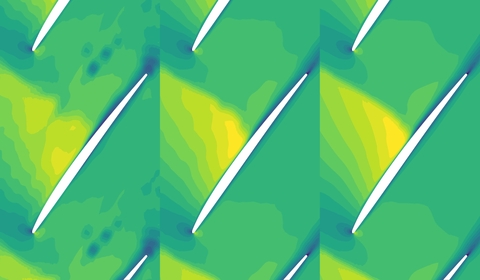}} 
}
\caption{Predictions using the backbone trained solo with only 1\% (left), 5\% (middle) of the dataset versus ground-truth (right)
of variable $V_x$ for \texttt{comp-rotor} near tip ($r_{S_1}=0.8$).}
\label{fig:backbone_solo_comp_rotor_1pc}
\end{center}
\vskip -0.15in
\end{figure}

To verify that the interpolation to unseen coordinates is smooth, we conduct the following experiment:
we select only 1\% and 5\% of the samples of the dataset $\mathcal{D}^{b-net}_\text{comp-rotor}$ for training.
In Figure \ref{fig:backbone_solo_comp_rotor_1pc}, we see that the flow solution is nevertheless
not erratic and flow characteristics are well reproduced.
As expected, the quality is not as good as in Figure \ref{fig:backbone_solo_comp_rotor} (bottom row).
In this experiment, we clearly have too few data points. Our model was implicitly biased
to smooth solutions by keeping the batch size the same as in the experiment with all data points.
With only 5\% of the samples, results are already very good.

\subsection{Hyper-Net}

The number of parameters of the hyper-net is generally huge,
proportional to the number of parameters of the backbone-net,
the constant of proportionality being the hidden dimension of the hyper-net.
This explains why we took great care to have a small but nevertheless accurate backbone-net.
By applying  shift modulation (see Section~\ref{The Hyper-Net}),
the number of trainable parameters was reduced to roughly 0.5Mio parameters.

To better assess the generalization performance of the hyper-net in the low data regime,
we consider circa 130 training samples. 
The training set is divided into training and validation sets (85/15 split).

Following \cite{lee2019self}, we consider experimentally two sources of uncertainty:
1/ the fold for training,
and 2/ the initialization of the parameters.
In Table \ref{table:hyper-summary-comp} and \ref{table:hyper-summary-turb},
we report the mean and standard deviation of the losses
calculated over 20 trials using different random seeds and training folds
(generated by shuffling the dataset prior to the split).
Notice that the hyperparameters for regularizing the model
were optimized for a single random split prior to the trials and then frozen.
We observe that the model performance is the same across all folds,
so we are confident that the observations made henceforth for a random split
will hold for any split, in particular the agreement
between true and predicted quantities of interest is remarkable, with 
correlation coefficients close to 1, Figure \ref{fig:hyper_comp_corr_stator},
\ref{fig:hyper_comp_corr_rotor}, \ref{fig:hyper_turb_corr_stator} and \ref{fig:hyper_turb_corr_rotor}.

\begin{figure}[!ht]
\begin{center}
\scalebox{1}[1.3]{
\centerline{\includegraphics[width=\columnwidth, trim=0.cm 0.3cm 0.4cm 0.8cm,clip]{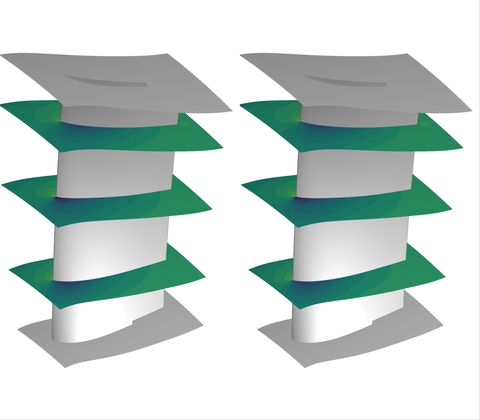}} 
}
\vskip -0.35in
\caption{Predictions (left) and ground-truth (right) of the pressure for sample 301 of the dataset $\mathcal{D}^{h-net}_\text{comp-stator}$
at $r_{S_1}=0.2$ (close to the hub), $r_{S_1}=0.5$ (midspan) and $r_{S_1}=0.8$ (close to the tip).
This test configuration has a loss close to the average test loss.}
\label{fig:hypernet_comp_stator_full}
\end{center}
\end{figure}

\begin{figure}[!ht]
\begin{center}
\scalebox{1}[1.2]{
\centerline{\includegraphics[width=\columnwidth, trim=0.cm 0.3cm 0.cm 0.cm,clip]{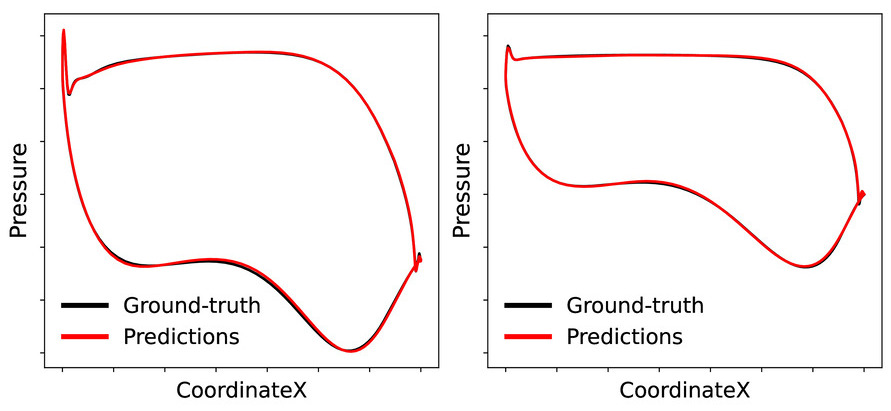}} 
}
\vskip 0.in
\caption{Pressure profiles at $r_{S_1}=0.2$ (left) and $r_{S_1}=0.8$ (right) for sample 153 of the dataset $\mathcal{D}^{h-net}_\text{turb-stator}$.
This test configuration has a loss close to the the average test loss.}
\label{fig:hypernet_turb_stator_full}
\end{center}
\vskip -0.15in
\end{figure}

To illustrate that model predictions are at least visually satisfactory,
we pick up a test sample with a loss close to the average test loss from the dataset $\mathcal{D}^{h-net}_\text{comp-stator}$
and plot the pressure field, Figure \ref{fig:hypernet_comp_stator_full}.
From that perspective, predictions are excellent.
We proceed similarly with the dataset $\mathcal{D}^{h-net}_\text{turb-stator}$
and plot the pressure profiles
since they are of uttermost importance for turbomachine designers, Figure \ref{fig:hypernet_turb_stator_full}.

\begin{figure}[!ht]
\begin{center}
\scalebox{1}[1]{
\centerline{\includegraphics[width=\columnwidth, trim=0.cm 0.cm 0.cm 0.cm,clip]{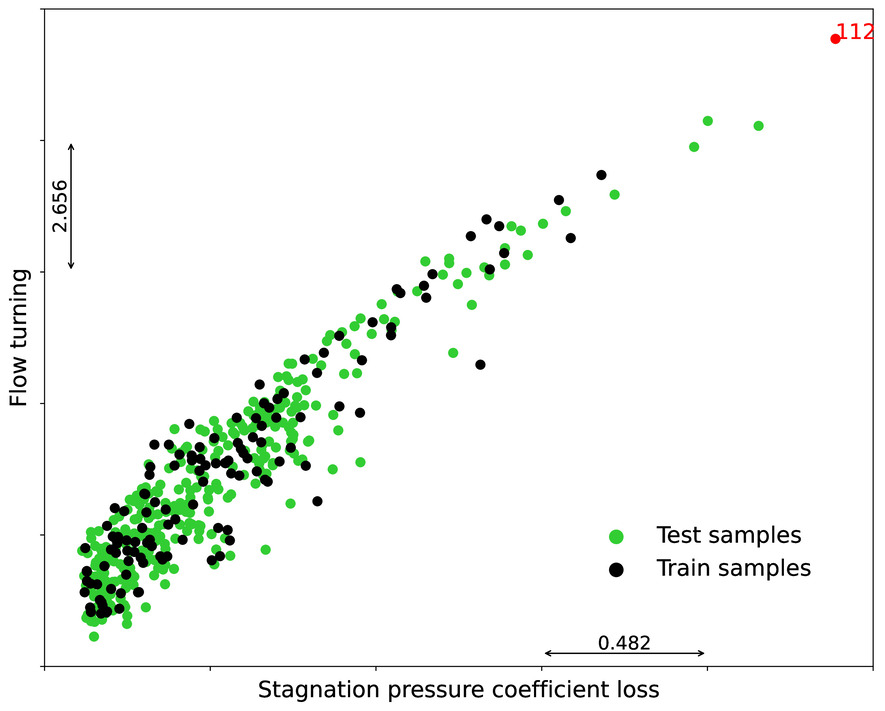}} 
}
\vskip -0.1in
\caption{Aerodynamic characteristics of the samples randomly generated from the blade \texttt{comp-rotor}.}
\label{fig:hypernet_comp_rotor_characteristics_sample_112}
\end{center}
\end{figure}

Now we specifically focus on the dataset $\mathcal{D}^{h-net}_\text{comp-rotor}$
randomly build around the low-pressure compressor rotor blade \texttt{comp-rotor}:
it is the most challenging from all datasets we tested our method on 
since the flow of many configurations in this dataset
is highly inhomogeneous radially with a large separation induced by a strong shock wave
from midspan outwards.
Those are mostly the configurations on the right in Figure~\ref{fig:hypernet_comp_rotor_characteristics_sample_112}
because their aerodynamic loss is significantly high.
What makes \texttt{comp-rotor}, and so all configurations in this dataset, even more particular
is that the blade is modelled with a tip clearance; the blade
rotates while the casing is non-rotating.
This raises severe difficulty in accurately predicting the flow solution.

\begin{figure}[!ht]
\scalebox{0.75}[0.75]{
  \centering
  \subfloat{\includegraphics[width=0.45\columnwidth]{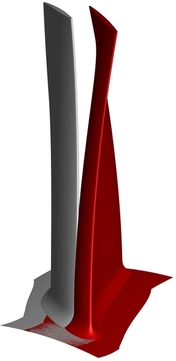}}  
}
  \hfill
\scalebox{0.75}[0.75]{
  \centering
  \subfloat{\includegraphics[width=0.45\columnwidth]{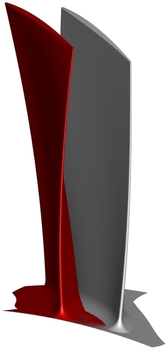}}
}
  \caption{Geometry of sample 112 in red and sample 465 in grey.}
  \label{fig:geometries}
\end{figure}

From the losses reported in Table~\ref{table:hyper-summary-comp} for both compressor datasets,
it is clear that our model overfits the training dataset
despite the use of dropout to mitigate overfitting.
Notice that our model has a small memory footprint:
It does not exceed 12GB
even though all samples are read in at the start to avoid IO intensive operations during training.
To illustrate the predictive capability of our model,
we consider test sample 112
that clearly lies outside the support
of training samples in the dataset, 
see Figure \ref{fig:hypernet_comp_rotor_characteristics_sample_112}.
The geometry of this sample is unconventional  and
definitively not interesting for designers because of the high aerodynamics losses it generates,
but to stress test our model it is the perfect candidate.
Its geometry is compared to a more conventional blade in Figure \ref{fig:geometries}.
Intuitively, we expect that our model will have trouble with new samples that lie outside the convex hull of the
training dataset as sample 112. But the plot is deceptive since it is low-dimensional:
It cannot fully convey a sense about how the convex hull of the dataset looks like.

\begin{figure}[!ht]
\begin{center}
\scalebox{1}[1.2]{
\centerline{\includegraphics[width=\columnwidth, trim=0.cm 0.3cm 0.cm 0.cm,clip]{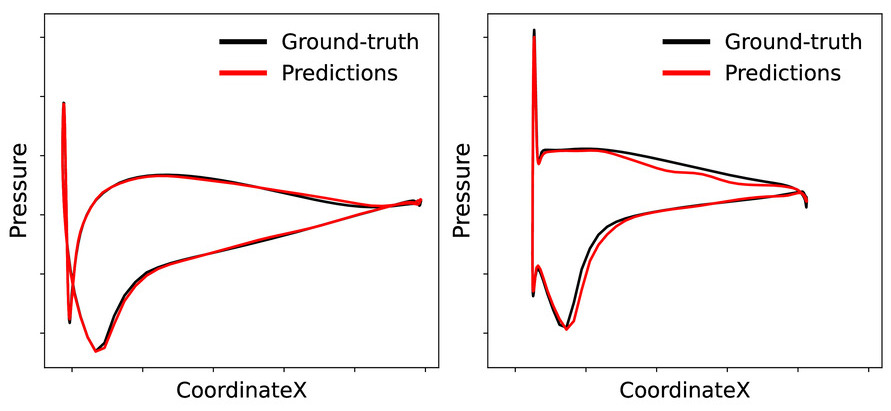}} 
}
\vskip 0.in
\caption{Pressure profiles at $r_{S_1}=0.2$ (left) and $r_{S_1}=0.8$ (right) for sample 112 of the dataset $\mathcal{D}^{h-net}_\text{comp-rotor}$.}
\label{fig:hypernet_comp_rotor_112_pressure_profile}
\end{center}
\end{figure}

\begin{figure}[!ht]
\begin{center}
\scalebox{1}[1]{
\centerline{\includegraphics[width=\columnwidth, trim=0.cm 0.3cm 0.cm 1.2cm,clip]{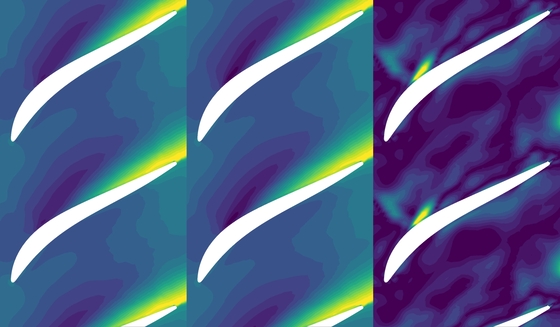}} 
}
\vskip 0.in
\caption{Prediction (left), ground-truth (middle) and absolute differences (right) for the radial velocity component $V_r$ for sample 112 at $r_{S_1}=0.2$ (near hub).}
\label{fig:hyper_comp_rotor_hub_Vr}
\end{center}
\end{figure}

\begin{figure}[!ht]
\begin{center}
\scalebox{1}[1]{
\centerline{\includegraphics[width=\columnwidth, trim=0.cm 0.3cm 0.cm 1.2cm,clip]{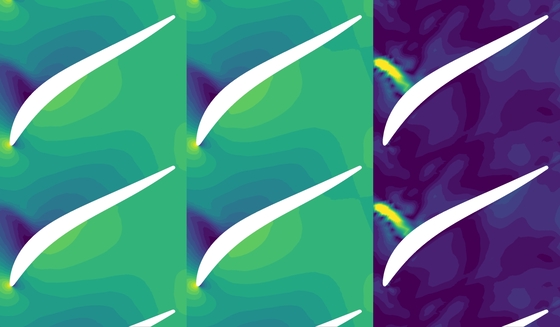}} 
}
\vskip 0.in
\caption{Prediction (left), ground-truth (middle) and absolute differences (right) for the pressure $p$ for sample 112 at $r_{S_1}=0.2$ (near hub).}
\label{fig:hyper_comp_rotor_hub_p}
\end{center}
\end{figure}

\begin{figure}[!ht]
\begin{center}
\scalebox{1}[1]{
\centerline{\includegraphics[width=\columnwidth, trim=0.cm 0.3cm 0.cm 1.2cm,clip]{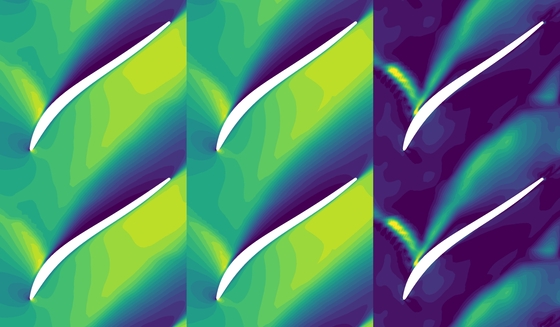}} 
}
\vskip 0.in
\caption{Prediction (left), ground-truth (middle) and absolute differences (right) for the axial velocity $V_x$ for sample 112 at $r_{S_1}=0.5$ (midpsan).}
\label{fig:hyper_comp_rotor_midspan_Vx}
\end{center}
\end{figure}

\begin{figure}[!ht]
\begin{center}
\scalebox{1}[1]{
\centerline{\includegraphics[width=\columnwidth, trim=0.cm 0.3cm 0.cm 1.2cm,clip]{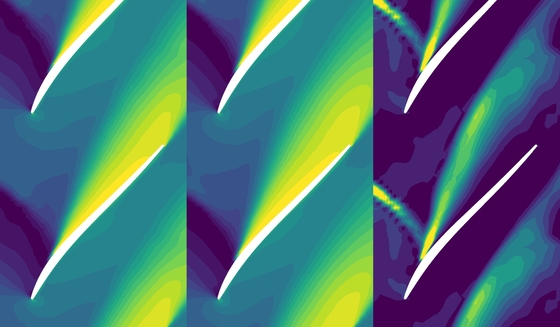}} 
}
\vskip 0.in
\caption{Prediction (left), ground-truth (middle) and absolute differences (right) for the tangential velocity $V_\theta$ for sample 112 at $r_{S_1}=0.8$ (near tip).}
\label{fig:hyper_comp_rotor_tip_Vtheta}
\end{center}
\end{figure}

Figures \ref{fig:hyper_comp_rotor_hub_Vr}-\ref{fig:hyper_comp_rotor_tip_Vtheta}
show the excellent performance of the hyper-net
in predicting challenging 3D flow solutions
of blades with complex geometry.
We can nevertheless observe some minor discrepancies, in particular 
the predicted show wave from midspan outwards seems to be slightly more smeared out, 
not as sharply captured as in the CFD simulation.
Figure \ref{fig:hypernet_comp_rotor_112_pressure_profile} depicts the pressure distribution
around the blade. Close to the hub, the agreement between ground-truth and predictions is particularly good.
Close to the tip, our model has trouble with the large flow separation on the suction side
and with the attached flow on the pressure side.
A domain where such a proxy can shine is blade shape optimization because it is cheap and accurate.
However, until now we only provided a visual assessment of the model prediction ability.
To be useful as a low-fidelity model in a multi-fidelity optimization, a model has to predict
quantities of interest also accurately, by far a much harder task.
This ability is usually demonstrated with correlation plots:
fidelity models should be well correlated, with a correlation coefficient close to 1 \cite{toal2015some}.
Figure \ref{fig:hyper_comp_corr_rotor} reveals that the hyper-net can almost perfectly predict massflow and flow turning.
Predicting aerodynamic losses is usually more difficult \cite{bonnet2022airfrans,kalaydjian2023packed}
since it is a highly non-linear function of the predictions of the primitive variables,
but the hyper-net tops here with a Pearson correlation coefficient as high as 0.995 on the test set.

\section{Conclusion}

In this work, we derived a fast and accurate proxy to CFD solvers based on an implicit neural representation.
Our model comprises a backbone-net that
establishes the mapping of a point in the computational domain to
its flow features. It is instantiated as a small coordinate-based MLP.
We conditioned the weights of the backbone-net onto the
blade geometry using a hyper-net.
It is instantiated as a small residual network.
By doing so, we could directly predict the full 3D flow solution
for any unseen blade geometry.
We showed that our model can make accurate predictions of full 3D flows
of compressor blades. Our model succeeded even though the flow exhibits
large separations induced by strong shock waves and is inhomogeneous from hub to tip.
Furthermore, for quantities-of-interest like massflow, flow turning and efficiency
aggregated from the model predictions, our model has also very good predictive capabilities.
As a result, our model is likely to be useful in future
as a low-fidelity model in a multi-fidelity optimization
of blade shapes.

\section{Acknowledgments}

The authors would like to thank MTU Aero Engines for the permission to publish this paper.
The research work associated with this publication has been supported
by the German Federal Ministry for Economic Affairs and Climate Action
under grant number 20X1909E.
The funding of the work through the 1st call of the Federal Aviation Research Program VI (LuFo VI-1),
grant project title ‘DIGIfly’, is gratefully acknowledged.
The authors are responsible for the content of this publication.

\bibliography{paper}
\bibliographystyle{icml2021}

\appendix
\onecolumn
\section{Compressor stator and rotor}

\makeatletter
\setlength{\@fptop}{0pt}
\makeatother

\vskip -0.2in
\begin{table}[!ht]
\caption{Summary of hyper-net training on the compressor dataset. A dataset is a set of simulations.}
\label{table:hyper-summary-comp}
\vskip +0.1in
\begin{center}
\begin{small}
\begin{sc}
\begin{tabular}{lrrr}
\toprule
Dataset & $\mathcal{D}^{h-net}_\text{comp-stator}$ & $\mathcal{D}^{h-net}_\text{comp-rotor}$ \\
\midrule
Dim. design space       &   13                &  6                       \\
\#Surface triangles/K   &   61                &  42                     \\
\#Surface nodes/K       &   31                &  21                     \\
\#Training samples      & 135                 & 128         \\
\#Test samples          & 264                 & 316         \\
\#Epochs                & 400                 & 400         \\
Training loss           & 1.9e-3 $\pm$ 5.7e-5 & 3.3e-3 $\pm$ 7.8e-5      \\
Validation loss         & 3.7e-3 $\pm$ 1.3e-3 & 4.2e-3 $\pm$ 3.5e-4      \\
Test loss               & 3.8e-3 $\pm$ 1.5e-3 & 4.3e-3 $\pm$ 1.3e-4      \\
Training time/hours     & 23                  & 18          \\
Peak memory/GB          & 14                  & 12          \\                 
\bottomrule
\end{tabular}
\end{sc}
\end{small}
\end{center}
\vskip -0.1in
\end{table}

\clearpage 

\begin{figure*}[!h]
\begin{center}
\scalebox{0.6}[0.6]{
\centerline{\includegraphics[width=\textwidth, trim=0.cm 0.cm 0.cm 0.cm,clip]{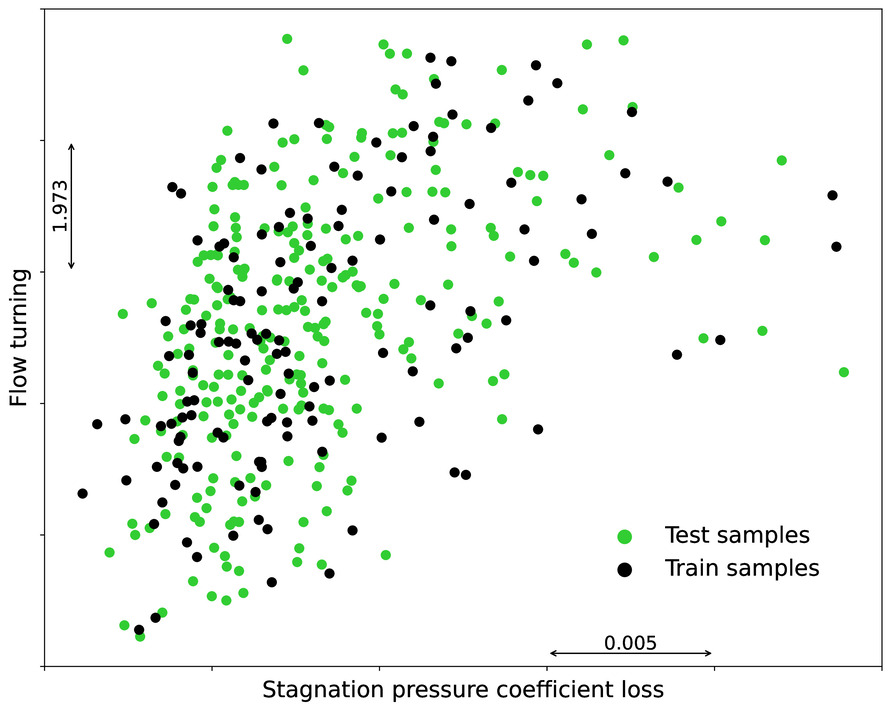}} 
}
\caption{Aerodynamic characteristics of the samples randomly generated from the blade \texttt{comp-stator}.}
\label{fig:hypernet_comp_stator_characteristics}
\end{center}
\end{figure*}

\begin{figure*}[!h]
\begin{center}
\scalebox{1}[1]{
\centerline{\includegraphics[width=\textwidth, trim=0.cm 0.cm 0.cm 0.cm,clip]{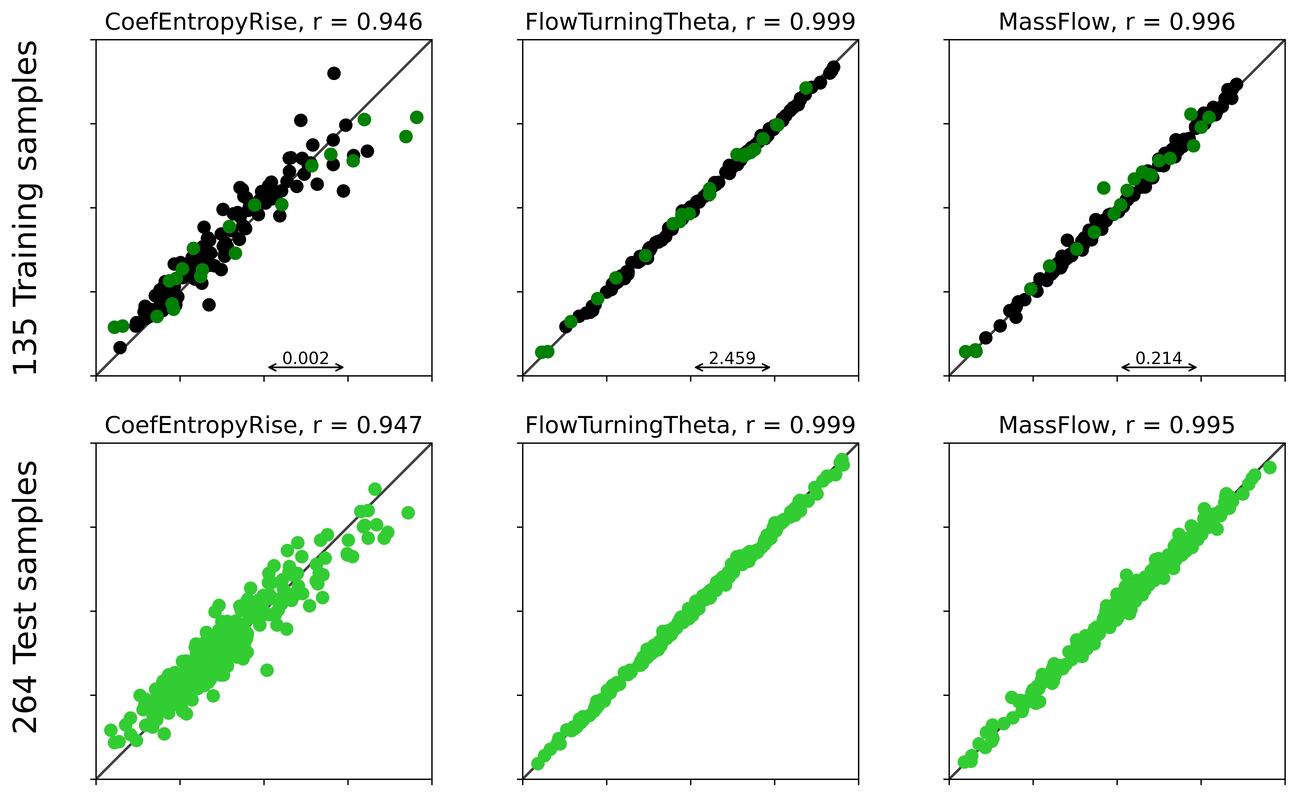}} 
}
\end{center}
\caption{Correlation plots for the samples randomly generated from the blade \texttt{comp-stator}.
$r$ designates the Pearson product-moment correlation coefficient.
In the top row of each figure, training samples are in black whereas validation samples are in green.}
\label{fig:hyper_comp_corr_stator}
\end{figure*}

\clearpage 

\begin{figure}[!h]
\begin{center}
\scalebox{0.6}[0.6]{
\centerline{\includegraphics[width=\columnwidth, trim=0.cm 0.cm 0.cm 0.cm,clip]{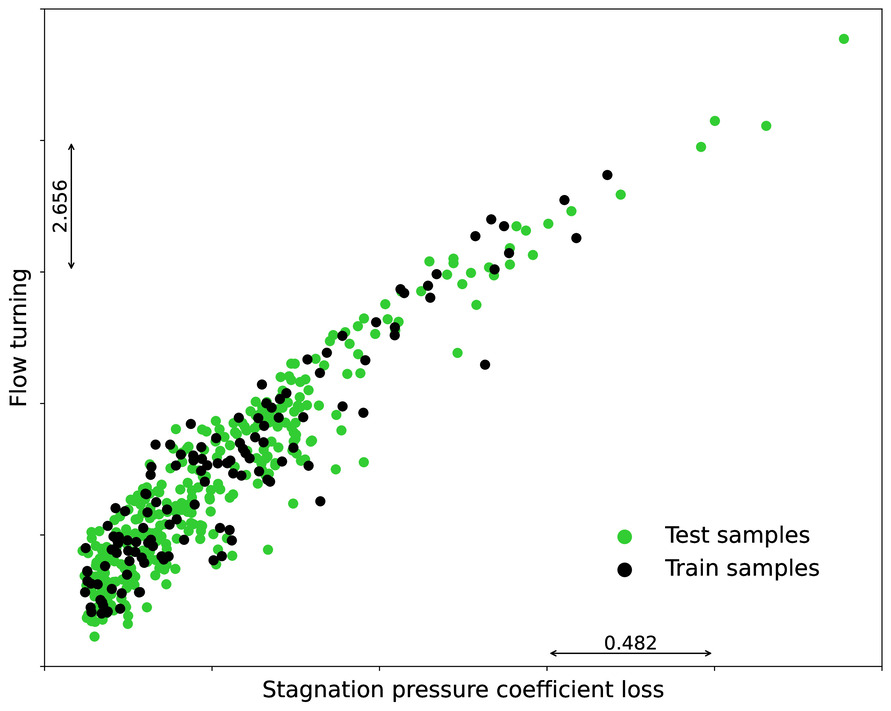}} 
}
\vskip -0.1in
\caption{Aerodynamic characteristics of the samples randomly generated from the blade \texttt{comp-rotor}.}
\label{fig:hypernet_comp_rotor_characteristics}
\end{center}
\end{figure}

\begin{figure*}[!h]
\begin{center}
\scalebox{1}[1]{
\centerline{\includegraphics[width=\textwidth, trim=0.cm 0.cm 0.cm 0.cm,clip]{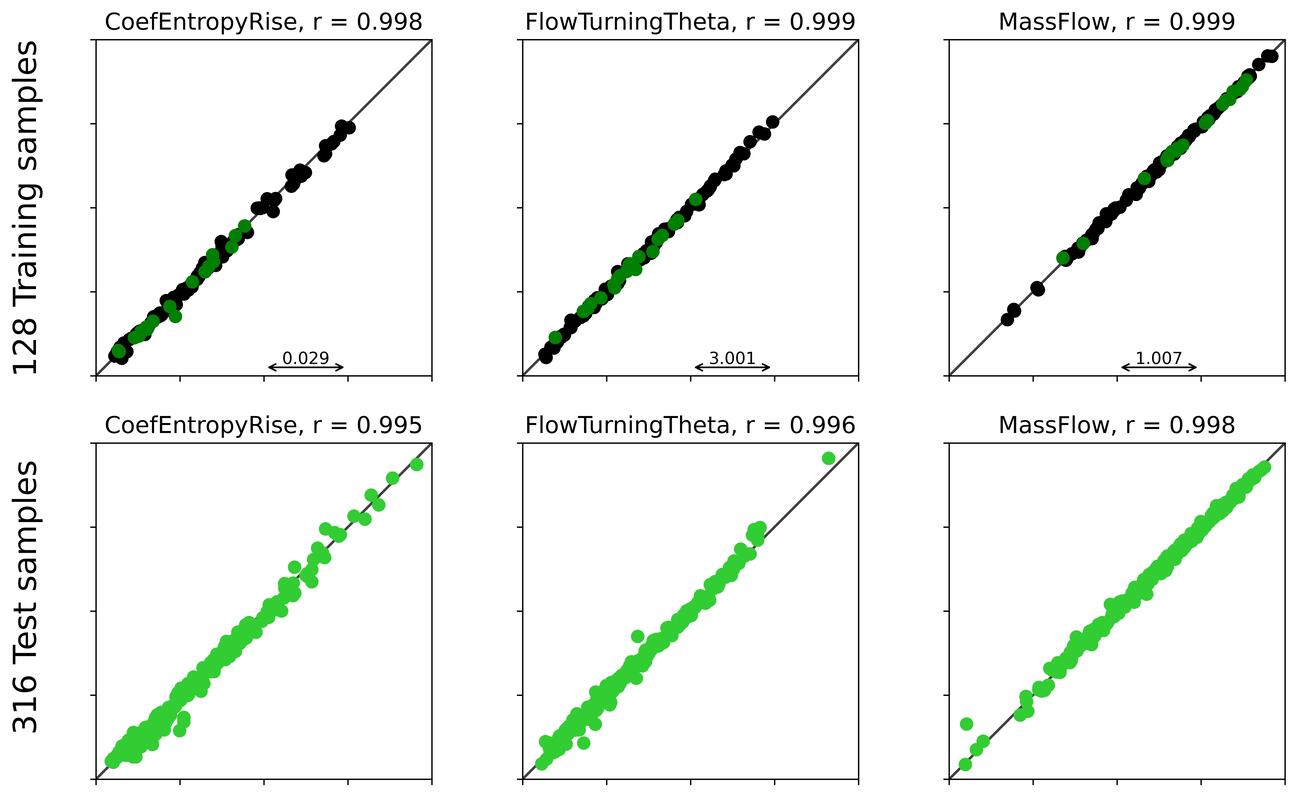}} 
}
\end{center}
\caption{Correlation plots for the samples randomly generated from the blade \texttt{comp-rotor}.
$r$ designates the Pearson product-moment correlation coefficient.
In the top row of each figure, training samples are in black whereas validation samples are in green.}
\label{fig:hyper_comp_corr_rotor}
\end{figure*}

\clearpage

\section{Turbine stator and rotor}

\vskip -0.2in
\begin{table}[!ht]
\caption{Summary of hyper-net training on the turbine dataset. A dataset is a set of simulations.}
\label{table:hyper-summary-turb}
\vskip +0.1in
\begin{center}
\begin{small}
\begin{sc}
\begin{tabular}{lrrr}
\toprule
Dataset & $\mathcal{D}^{h-net}_\text{turb-stator}$ & $\mathcal{D}^{h-net}_\text{turb-rotor}$ \\
\midrule
Dim. design space       &   17                 &  13                   \\
\#Surface triangles/K   &   42                 &  42                  \\
\#Surface nodes/K       &   21                 &  21                  \\
\#Training samples      & 132                  & 129                   \\
\#Test samples          & 497                  & 459                   \\
\#Epochs                & 400                  & 400                   \\
Training loss           & 2.4e-3 $\pm$ 3.2e-5  & 2.7e-3 $\pm$ 3.1e-5   \\
Validation loss         & 3.3e-3 $\pm$ 2.8e-4  & 3.7e-3 $\pm$ 2.3e-4   \\
Test loss               & 3.3e-3 $\pm$ 8.4e-5  & 3.7e-3 $\pm$ 1.3e-4   \\
Training time/hours     & 18                   & 18                    \\
Peak memory/GB          & 18                   & 18                    \\                 
\bottomrule
\end{tabular}
\end{sc}
\end{small}
\end{center}
\vskip -0.1in
\end{table}

\clearpage 

\begin{figure*}[!h]
\begin{center}
\scalebox{0.6}[0.6]{
\centerline{\includegraphics[width=\textwidth, trim=0.cm 0.cm 0.cm 0.cm,clip]{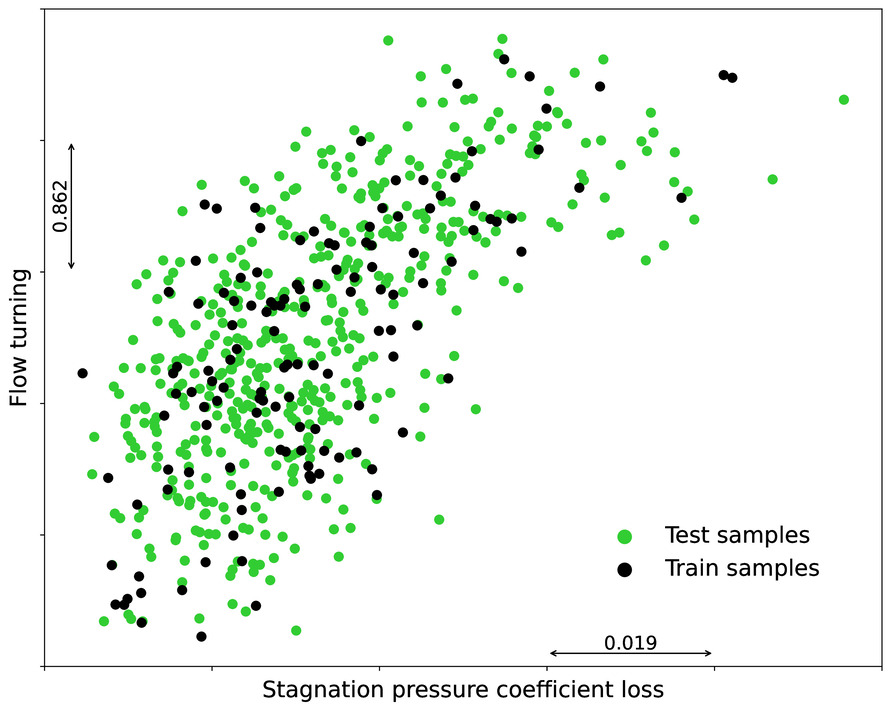}} 
}
\caption{Aerodynamic characteristics of the samples randomly generated from the blade \texttt{turb-stator}.}
\label{fig:hypernet_turb_stator_characteristics}
\end{center}
\end{figure*}

\begin{figure*}[!h]
\begin{center}
\scalebox{1}[1]{
\centerline{\includegraphics[width=\textwidth, trim=0.cm 0.cm 0.cm 0.cm,clip]{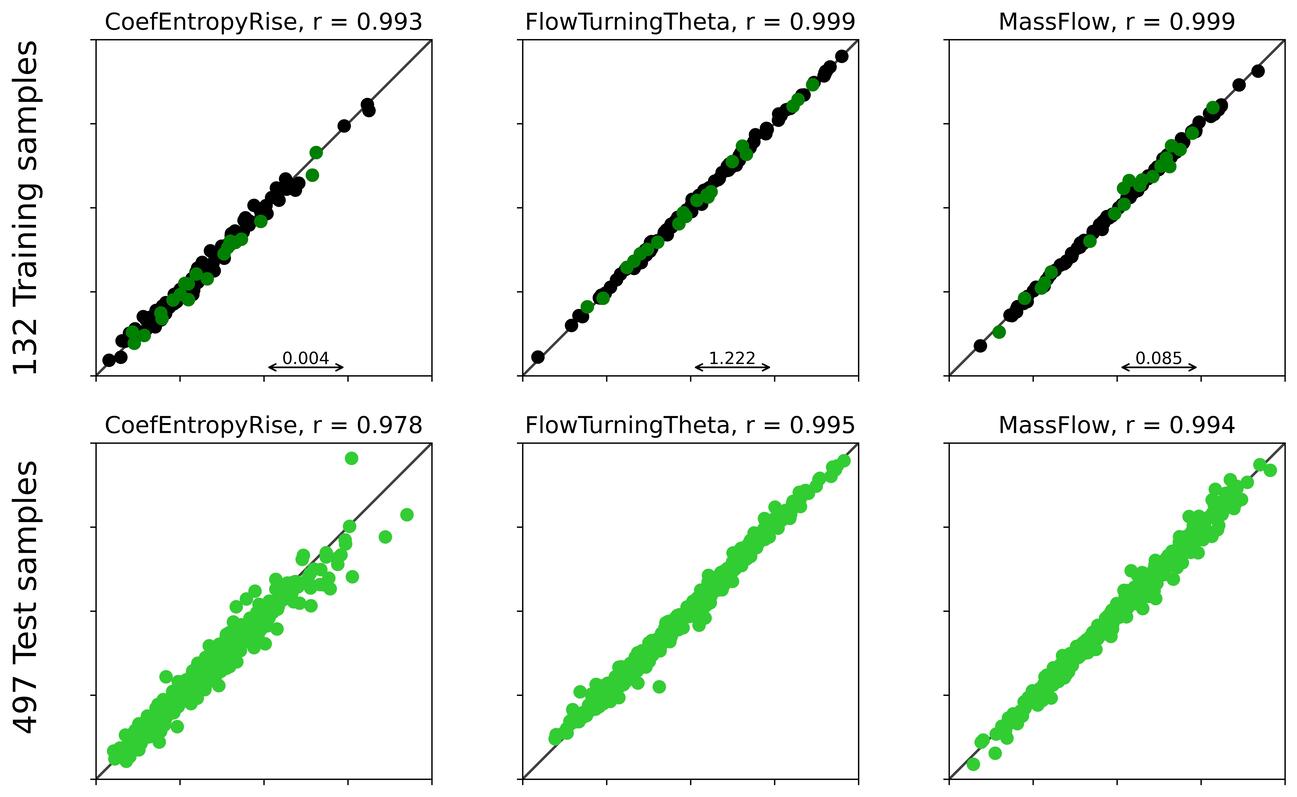}} 
}
\end{center}
\caption{Correlation plots for the samples randomly generated from the blade \texttt{turb-stator}.
$r$ designates the Pearson product-moment correlation coefficient.
In the top row of each figure, training samples are in black whereas validation samples are in green.}
\label{fig:hyper_turb_corr_stator}
\end{figure*}

\clearpage 

\begin{figure}[!h]
\begin{center}
\scalebox{0.6}[0.6]{
\centerline{\includegraphics[width=\columnwidth, trim=0.cm 0.cm 0.cm 0.cm,clip]{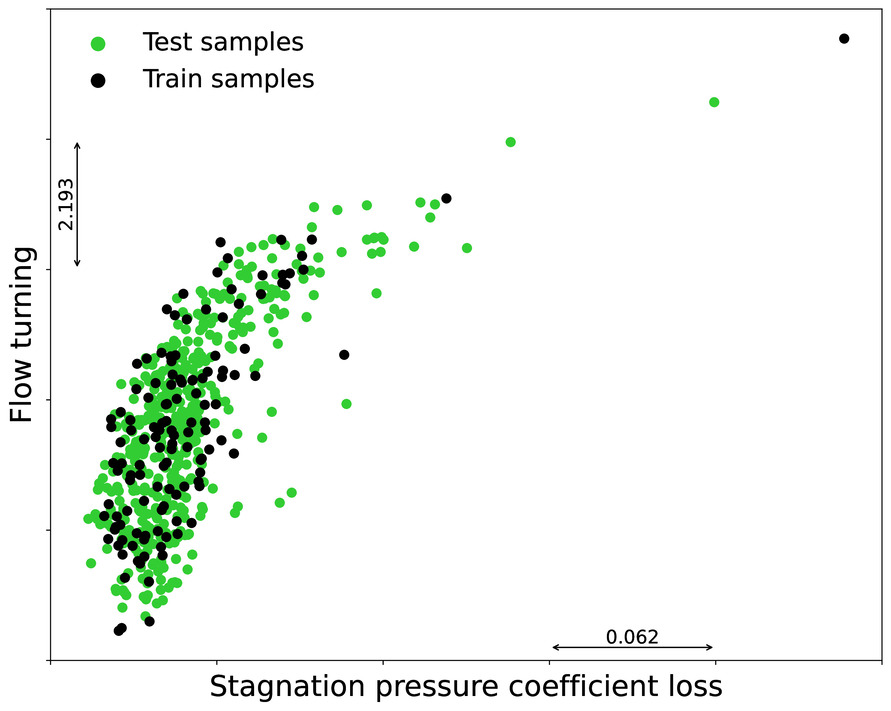}} 
}
\vskip -0.1in
\caption{Aerodynamic characteristics of the samples randomly generated from the blade \texttt{turb-rotor}.}
\label{fig:hypernet_turb_rotor_characteristics}
\end{center}
\end{figure}

\begin{figure*}[!h]
\begin{center}
\scalebox{1}[1]{
\centerline{\includegraphics[width=\textwidth, trim=0.cm 0.cm 0.cm 0.cm,clip]{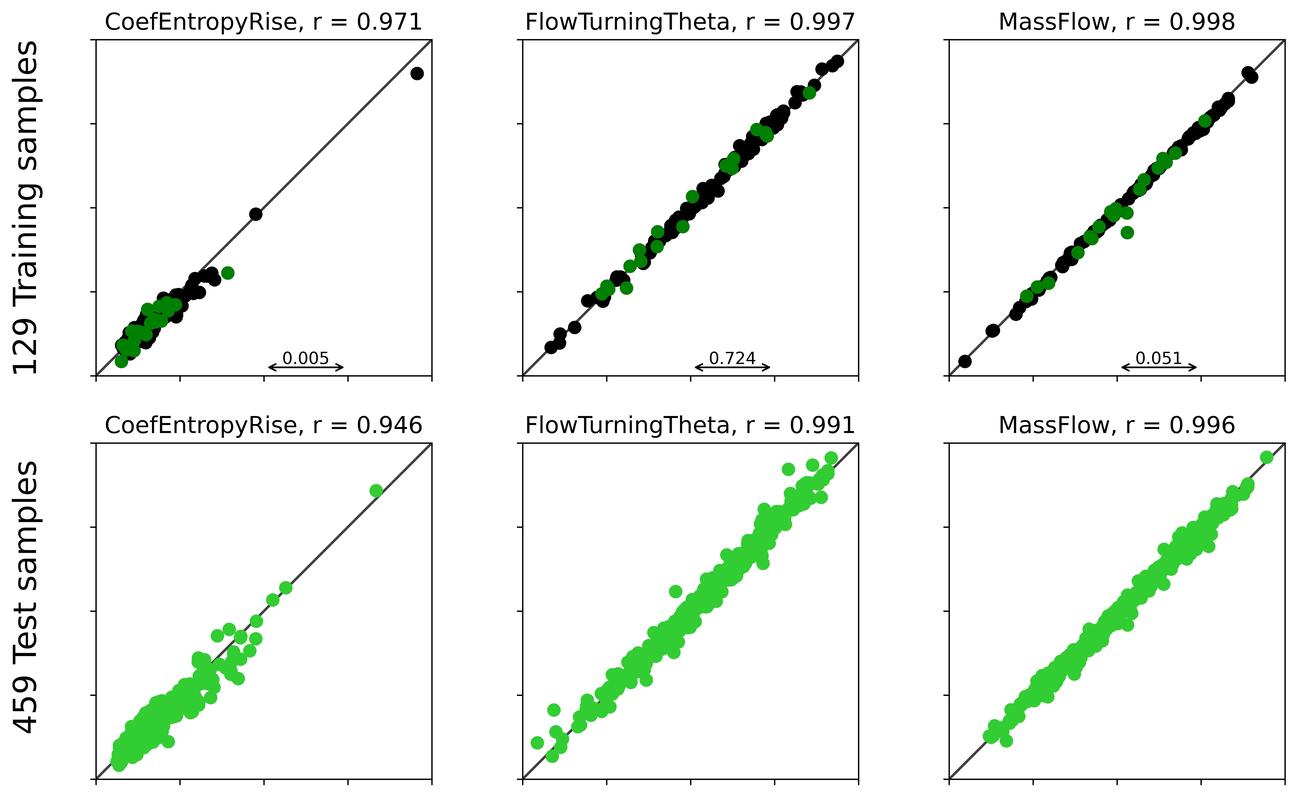}} 
}
\end{center}
\caption{Correlation plots for the samples randomly generated from the blade \texttt{turb-rotor}.
$r$ designates the Pearson product-moment correlation coefficient.
In the top row of each figure, training samples are in black whereas validation samples are in green.}
\label{fig:hyper_turb_corr_rotor}
\end{figure*}

\clearpage

\section{Embedding Dimension}

\begin{figure*}[!h]
\begin{center}
\scalebox{0.6}[0.6]{
\centerline{\includegraphics[width=\textwidth, trim=0.cm 0.cm 0.cm 0.cm,clip]{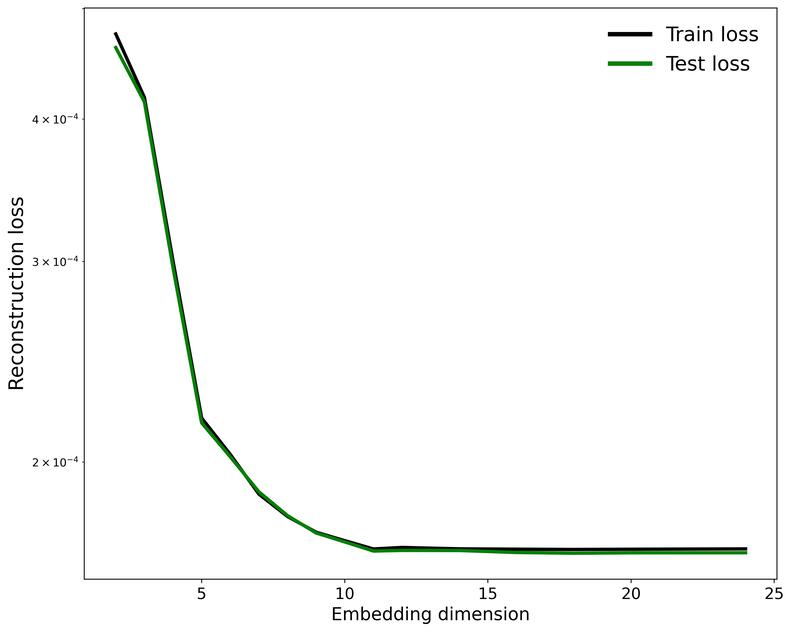}} 
}
\end{center}
\caption{Reconstruction loss for the samples in \texttt{turb-stator}.
The blade parameterization has 17 degrees of freedom.}
\label{fig:turb_stator_geo}
\end{figure*}

The dimension of the embedding for a given dataset has to be set greater
than the intrinsic dimension of the dataset that is unknown. However, it is less than or equal to 
the number of degrees of freedom of the blade parameterization that is known.
In our experiments it is usually less than 16 except for the dataset \texttt{turb-stator} for
which it is 17. So we carried out a study to find out a reasonable minimum embedding dimension
as follows: From the blade embeddings, the blades are reconstructed
and the quality of the reconstruction is assessed by the Chamfer loss.
To generate a point cloud with variable number of points,
we sample a vector of dimension 3 from the normal distribution, append it to the blade embedding
and push it through a small MLP that outputs the spatial coordinates.
Figure \ref{fig:turb_stator_geo} shows that an embedding dimension of 12 is sufficient
for a good reconstruction.

\end{document}